\documentclass[10pt,twocolumn,letterpaper]{article}

\usepackage[pagenumbers]{cvpr}      

\usepackage{graphicx}
\usepackage{amsmath}
\usepackage{amssymb}
\usepackage{booktabs}
\usepackage{graphicx}
\usepackage{bm}
\usepackage{algorithm}
\usepackage{algpseudocode}
\usepackage{mathtools}
\usepackage{amsthm}
\usepackage{booktabs}
\usepackage{siunitx}
\usepackage{comment}
\usepackage{cite}
\usepackage[pagebackref,breaklinks,colorlinks]{hyperref}
\usepackage[T1]{fontenc}
\usepackage[utf8]{inputenc}

\usepackage{xr-hyper}

\makeatletter
\newcommand*{\addFileDependency}[1]{
  \typeout{(#1)}
  \@addtofilelist{#1}
  \IfFileExists{#1}{}{\typeout{No file #1.}}
}
\makeatother

\hypersetup{allcolors=red}

\usepackage[capitalize]{cleveref}
\crefname{section}{Sec.}{Secs.}
\Crefname{section}{Section}{Sections}
\Crefname{table}{Table}{Tables}
\crefname{table}{Tab.}{Tabs.}

\def\cvprPaperID{*****} 
\def\confName{CVPR}
\def\confYear{2026}

\algrenewcommand\algorithmicrequire{\textbf{Input:}}
\algrenewcommand\algorithmicensure{\textbf{Output:}}

\begin{document}

\title{Real-Time Drone Detection in Event Cameras via Per-Pixel Frequency Analysis}

\author{Michael Bezick\\
Johns Hopkins University APL\\
{\tt\small michael.bezick@jhuapl.edu}
\and
Majid Sahin\\
Johns Hopkins University APL\\
{\tt\small majid.sahin@jhuapl.edu}
}
\maketitle

\begin{figure*}
    \centering
    \includegraphics[width=0.8\linewidth]{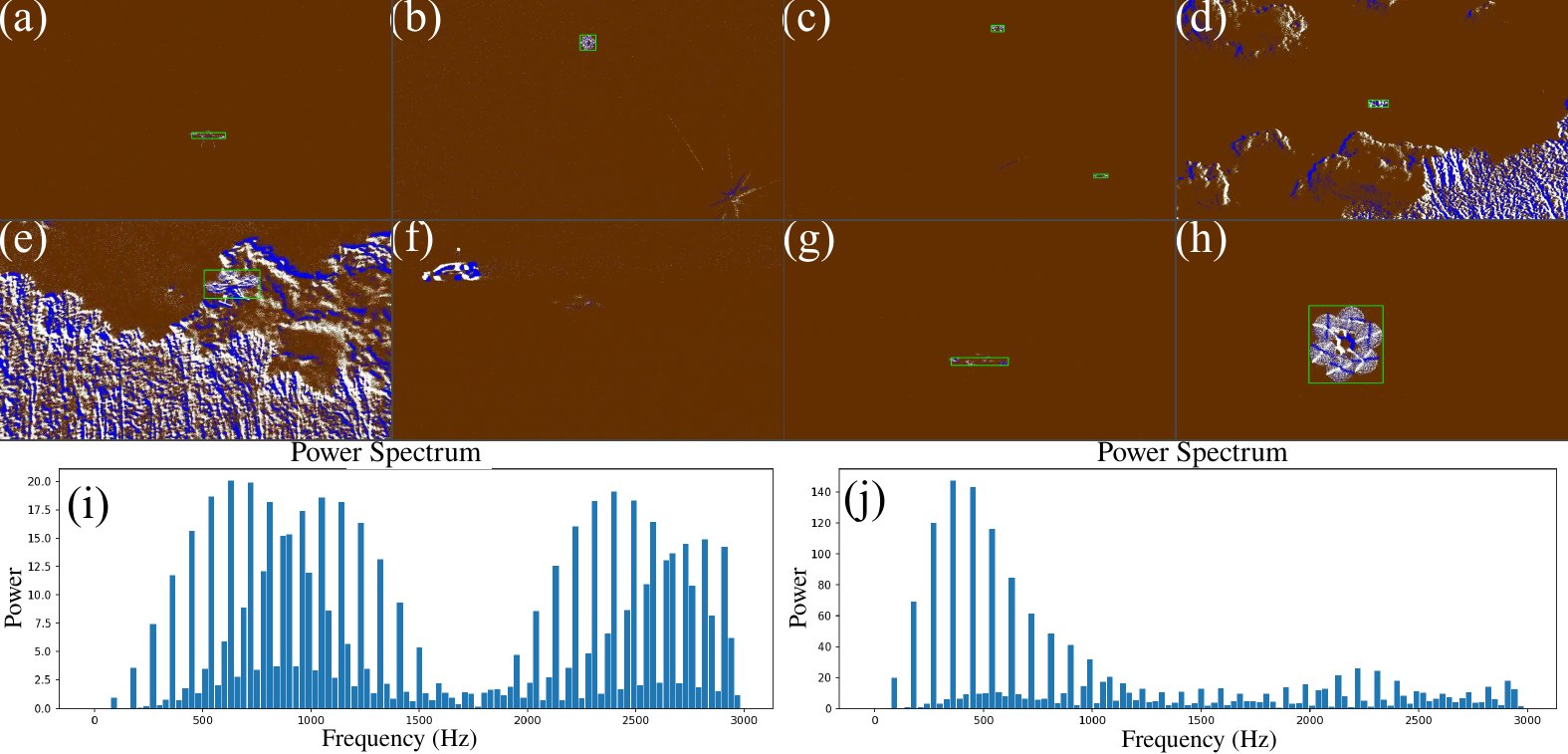}
    \caption{\textbf{(a)} Far distance drone test at 300m using the Navitar 100mm F2.8 lens. \textbf{(b)} Direct sunlight test demonstrating the wide dynamic range of event based vision. \textbf{(c)} Multi-drone test - both identified correctly. \textbf{(d)-(e)} Handheld shaky-cam examples. Algorithm is able to accurately isolate the drone from the background, while keeping latency to real-time speeds despite large event throughput. \textbf{(f)-(h)} Angle of elevation experiment at $0^\circ$, $10^\circ$, and $90^\circ$ respectively. At $0^\circ$, drone is barely visible (due to pose and possibly the focus of lens) and not identified, but at $10^\circ$, drone is accurately identified, showing a limitation of DDHF. At $90^\circ$ angle of elevation, the drone is most easily identified. \textbf{(i)} Power spectrum of pixel where two overlapping rotor paths projected onto. \textbf{(j)} Typical power spectrum of a single-rotor pixel.}
    \label{fig:main}
\end{figure*}

\begin{abstract}
    Detecting fast-moving objects, such as unmanned aerial vehicle (UAV), from event camera data is challenging due to the sparse, asynchronous nature of the input. Traditional Discrete Fourier Transforms (DFT) are effective at identifying periodic signals, such as spinning rotors, but they assume uniformly sampled data, which event cameras do not provide. We propose a novel per-pixel temporal analysis framework using the Non-uniform Discrete Fourier Transform (NDFT), which we call Drone Detection via Harmonic Fingerprinting (DDHF). Our method uses purely analytical techniques that identify the frequency signature of drone rotors, as characterized by frequency combs in their power spectra, enabling a tunable and generalizable algorithm that achieves accurate real-time localization of UAV.
    We compare against a YOLO detector under equivalent conditions, demonstrating improvement in accuracy and latency across a difficult array of drone speeds, distances, and scenarios. DDHF achieves an average localization F1 score of 90.89\%  and average latency of 2.39ms per frame, while YOLO achieves an F1 score of 66.74\% and requires 12.40ms per frame. Through utilization of purely analytic techniques, DDHF is quickly tuned on small data, easily interpretable, and achieves competitive accuracies and latencies to deep learning alternatives.
\end{abstract}


\section{Introduction}
\label{sec:intro}

Unmanned aerial vehicle (UAV) detection has become increasingly important in recent years due to the proliferation of consumer and commercial drones in sensitive airspace. Traditional machine learning-based detection pipelines often rely on large annotated datasets and compute-intensive training pipelines. These requirements limit scalability and make real-time deployment challenging, especially in resource-constrained environments. Moreover, frame-based RGB cameras struggle to capture the fine-grained, high-speed motion patterns of drone propellers under adverse conditions, such as low light, long range, or cluttered backgrounds, where motion blur and low contrast degrade performance.

Event cameras provide a fundamentally different sensing paradigm. Instead of sampling the entire image at fixed intervals, which may be redundant in unchanging scenes, they asynchronously record brightness changes at each pixel, achieving microsecond-level temporal resolution, high dynamic range (140 dB vs. 60 dB in standard CMOS sensors), and low power consumption \cite{gallego_event-based_2022}. These properties make event cameras particularly well-suited for detecting high-frequency motion signatures such as rotor oscillations, enabling robust operation across lightning conditions and environments where traditional sensors fail. However, processing event streams poses unique challenges. Unlike conventional video, event data are inherently sparse and irregular in time, rendering standard Fourier transform methods inapplicable. To exploit the rich temporal information of event cameras, spectral analysis must account for this non-uniformity without sacrificing temporal precision.

In this work, we propose a UAV harmonic fingerprinting framework for analyzing the power spectra, produced through the application of the Non-Uniform Discrete Fourier Transform (NDFT) to asynchronous event camera streams. Our method is highly parallelizable for real-time GPU processing, achieving real-time and highly accurate drone identification through solely analytical, non-neural methods.

Our main contributions are as follows:
\begin{enumerate}
    \item
    We formalize UAV rotor-induced harmonic structure in event streams and derive measurable spectral cues used for detection and identification.

    \item
    We show that event-to-spectrum construction used in  a prior frequency-mapping approach \cite{10.1007/978-3-031-92460-6_18} can be interpreted as an NDFT over asynchronous timestamps, leveraging this formulation to enable efficient GPU computation across time windows.

    \item
    We introduce a principled rotor fingerprinting method that localizes and distinguishes UAVs by identifying harmonic comb structure in their power spectra, yielding interpretable frequency-domain analysis.

    \item
    We build a GPU-accelerated detection pipeline that runs in real time and provide an in-depth evaluation of edge cases, highlighting robustness of event-based sensing compared to frame-based baselines.
\end{enumerate}

\section{Related Work}
\label{sec:related_work}
\noindent \textbf{Object Tracking with Event Cameras}. Mitrokhin et al. perform moving object tracking with event cameras by first fitting a parametric model to predict the motion of the camera, then iteratively identifying objects that do not move according to the model \cite{mitrokhin_event-based_2018}. Their method achieves faster-than-real-time computation speed and achieves mAP@50 scores between 0.85-0.92 on a variety of tasks. Mondal et al. present a graph spectral clustering technique to identify moving objects in event camera streams \cite{mondal_moving_2021}. They compare against achieving F1 scores at 0.75 intersection over union (IoU) of 0.67-0.91 on datasets of hands, cars \cite{almatrafi_distance_2020}, and streets \cite{hu_v2e_2021}, significantly improving in accuracy against density-based spatial clustering of applications with noise (DBSCAN), Meanshift \cite{chen_neuromorphic_2018}, and Gaussian mixture models \cite{piatkowska_spatiotemporal_2012} methods. However, these aforementioned methods are general motion trackers and are unable to distinguish between object types.
\\
\\
\noindent \textbf{Fourier Transforms on Event Camera Data.} Sabatier et al. introduce a method for performing the 2D spatial discrete Fourier transforms iteratively on each new arriving gray level event \cite{sabatier_asynchronous_2017}. The authors base their method on Stockham's algorithm \cite{cochran_what_1967}, which decomposes the transform into a matrix product of sparse matrices, which can be viewed as a network of several layers. To limit computation, if an incoming event doesn't induce a sufficient change in the signal from one layer to the next, their algorithm stores the information locally, waiting until the value of a node in a layer passes a threshold. The choice of threshold represents a tradeoff between accuracy and computation. The authors find that their method performs faster than equivalent frame-based methods. Grundmark et al. introduce a method for performing the temporal DFT on each pixel of an event camera stream \cite{grundmark_frequency_2023}. The authors define a uniformly-spaced grid and round events' timestamps to the nearest grid point. Instead of approximating events to the nearest grid point, Aitsam et al. operate directly on the timestamps of incoming events, which is equivalent to applying a NDFT to event polarities \cite{10.1007/978-3-031-92460-6_18}. The authors first conduct an initial experiment where they extract dominant frequencies of a spinning disc of known RPM, demonstrating high accuracies of their algorithm across lighting conditions, then they show their method can be applied to detect real-time vibrational anomalies in machinery.
\\\\
\noindent \textbf{Drone Detection.}
In recent years, there has been a significant increase in interest in detecting drones with the event camera sensing modality \cite{magrini2025dronedetectioneventcameras}. These approaches can be classified into frame based, time-aware (via point points, voxel grids, or time-retaining frames), end-to-end neuromorphic approaches with spiking neural networks and dedicated neuromorphic hardware. Mandula et al. accumulate frames over a short period of time, processing the data with a CNN detector \cite{10561168}. Eldeborg et al. utilize a Spiking Neural Network and an Artificial Neural Network on frame accumulations \cite{Eldeborg_Lundin_2025}. Stewart et al. utilize temporal histograms, identifying drone rotors by characteristic harmonic peaks with  a small neural network \cite{10.1145/3546790.3546800}. Sanket et al. introduce EVPropNet, where the authors train a deep convolutional neural network on a large-scale synthetic event camera dataset of drone propellers \cite{sanket2021evpropnetdetectingdronesfinding}.

Concerning traditional sensing modalities, Seidaliyeva et al. perform real-time optical drone detection by separating the task into identifying moving objects, then classifying detected objects \cite{seidaliyeva_real-time_2020}. The authors perform moving object detection through background subtraction on RGB images, which assumes a static camera, and then utilize a convolutional neural network (CNN) to perform classification. However, this method assumes a mobile drone, making stationary drones undetectable. Gl{\"u}ge et al. utilize CNNs, operating on spectrogram data, produced through Fourier transforms on radio frequency (RF) data \cite{gluge_robust_2024}. The authors achieve accuracies of approximately $80\%$, detecting drones as far as 670m.

\section{Method}
\label{sec:method}
\subsection{Event Cameras}
Different event cameras export different data modalities. Temporal contrast sensors report a discrete label corresponding to a positive or negative change in logarithmic illuminance, and temporal image sensors indicate instantaneous intensity. For this work, we focus on temporal contrast sensors. Contrast detection events are the tuple $(x, y, p, t)$, where $(x,y)$ are the pixel's coordinates, $p \in \{ -1, 1\}$ is the polarity, signifying a decrease / increase in illumination respectively, and $t$ is the timestep in microseconds. Temporal contrast sensors convert photocurrent into a logarithmically compressed voltage, and when this voltage exceeds or subceeds a threshold, a $+1$ or $-1$ polarity event is recorded for that pixel at the timestep it occurred. Then, the reference voltage for that pixel is reset to the new voltage, and the process repeats \cite{qin_event_2025}. Event cameras feature adjustable bias settings, such as temporal contrast event threshold, which determines how much the voltage change is required to register an event, low and high-pass filters that remove flickering noise and slow-changing elements respectively, and a refractory period that adjusts the sleep duration maintained by the sensor after each registered event \cite{dilmaghani_autobiasing_2024}.

\subsection{Rotor-Induced Frequency Signatures}

Rotating propellers modulate the brightness of the scene as blades periodically occlude and reveal the background. At a rotor pixel, the resulting event stream can be viewed as a non-uniform sampling of an underlying periodic rectangular wave of illuminance: during a short interval of each revolution the blade blocks the background, and for the remainder of the period the background is visible.

Let $f_{\mathrm{BPF}}$ denote the blade-pass frequency, $T_{\mathrm{BPF}} = 1 / f_{\mathrm{BPF}}$ its period, and let $\tau \in (0,1)$ be the duty cycle, i.e., the fraction of each period for which the pixel is occluded. An idealized occlusion signal at such a pixel can be written as the pulse train
\begin{equation}
    p(t)
    = \sum_{k=-\infty}^{\infty}
      \mathrm{rect}\!\left(
        \frac{t - k T_{\mathrm{BPF}}}{\tau T_{\mathrm{BPF}}}
      \right),
\end{equation}
where the rectangular function $\mathrm{rect}(x)$ is
\begin{equation}
    \mathrm{rect}(x) =
    \begin{cases}
        1,   & \text{if } |x| < \tfrac{1}{2}, \\
        0.5, & \text{if } |x| = \tfrac{1}{2}, \\
        0,   & \text{if } |x| > \tfrac{1}{2}.
    \end{cases}
\end{equation}

Because $p(t)$ is $T_{\mathrm{BPF}}$-periodic, it admits the Fourier series
\begin{equation}
    p(t)
    = \sum_{n=-\infty}^{\infty}
      c_n \, e^{i 2\pi n f_{\mathrm{BPF}} t},
    \qquad
    c_n = \tau\,\mathrm{sinc}(n \tau)\,e^{-i \pi n \tau},
\end{equation}
where $\mathrm{sinc}(x) = \frac{\sin(\pi x)}{\pi x}$. Taking the Fourier transform of this series yields a line spectrum
\begin{equation}
    P(f)
    = \sum_{n=-\infty}^{\infty}
      c_n \, \delta\!\bigl(f - n f_{\mathrm{BPF}}\bigr),
\end{equation}
which consists of impulses at integer multiples $f = n f_{\mathrm{BPF}}$ of the blade-pass frequency. Thus the power in the $n$-th harmonic satisfies
\begin{equation}
    |P(f)|^2 \;\propto\; |c_n|^2
    = \left(
        \frac{\sin(\pi n \tau)}{\pi n}
      \right)^2.
\end{equation}

In the UAV setting, this rotor-induced comb at $f_{\mathrm{BPF}}$ and its integer harmonics provides a distinctive fingerprint of propeller motion, enabling us to discriminate UAV pixels from background motion, turbulence, and other non-periodic clutter. Figure~\ref{fig:main}(i)–(j) shows typical examples of these harmonic comb spectra measured at individual rotor pixels in our data.

\subsection{Non-Uniform Fourier Transform}

Event cameras generate temporally sparse, polarity-encoded measurements in response to local brightness changes. Unlike conventional image sensors that acquire data at fixed frame intervals, event streams are asynchronous and non-uniformly sampled in time, with each pixel emitting events only when stimulated by motion or intensity variation. This sensing model naturally lends itself to frequency analysis: periodic motion, such as rotor blades, induces identifiable harmonic comb spectra, whereas background clutter and noise create near-uniform spectra.

Given a stream of $M$ events $\{(x, y,p_j, t_j) \}_{j=1}^M$ at pixel location $(x,y)$, following \cite{10.1007/978-3-031-92460-6_18} we compute the NDFT,

\begin{equation}
F_k = \sum_{j=1}^{M} p_j \exp \left( i k t_j \right),
\qquad k \in \mathcal{I}_N,
\end{equation}

with $\mathcal{I}_N$ the centered discrete frequency index set \cite{bagchi_nonuniform_1999}. The resulting complex spectrum $F_k$ captures periodic structure directly from asynchronous samples, without temporal binning or artificial resampling. We then compute the power spectrum $P_k = |F_k|^2$ to quantify frequency energy distribution. In this work, we set $\mathcal{I}_N = \{0, ..., N -1 \}$.

This formulation preserves the temporal precision of event sensors and avoids aliasing effects introduced by time quantization, as performed in~\cite{grundmark_frequency_2023}. Furthermore, the NDFT allows for evaluating the contribution of a single event to the power spectrum as it arrives, matching the asynchronous nature of event cameras; however, for GPU efficiency we choose to batch events. \footnote{Although NUFFTs~\cite{dutt_fast_1993} offer asymptotic acceleration for large sample counts, the event-camera setting typically produces tens of events per pixel per time window; thus, direct NDFT evaluation is faster than NUFFT variants due to reduced overhead.}

\subsection{Drone Classification on Frequency Responses}

\noindent\textbf{Spectral Flatness for Noise Rejection.} Event noise and background motion often produce uniform spectral responses, whereas rotor signals exhibit tone-like structure. We therefore first evaluate the spectral flatness (SF) of power spectra,

\begin{equation}
\text{SF} =
\frac{ \exp \left( \frac{1}{N} \sum_{k=1}^N \log P_k \right)}
{ \frac{1}{N} \sum_{k=1}^N P_k },
\end{equation}
where lower SF values indicate concentrated harmonic energy and high values correspond to uniform, noise-like spectra. We discard spectra where $\text{SF} > \tau_{\text{sf}}$, removing pixels lacking coherent periodic structure. This step reduces false positives while remaining insensitive to the absolute magnitudes of power spectra.
\\\\
\noindent\textbf{Median–Normalized Harmonic Comb for Rotor Signature Detection.}
Events generated from rotor blades produce a strong harmonic series (Figure~\ref{fig:main} (i)-(j)), characterized by a fundamental rotation frequency and its integer multiples. To detect such patterns, we compute a comb score for each candidate base frequency \(\omega\) by averaging local peak responses around the first \(M\) harmonics,
\begin{equation}
N(\omega) \;=\; \frac{1}{M_{\text{valid}}(\omega)} \sum_{m=1}^{M_{\text{valid}}(\omega)}
\max_{\,|k-m\omega|\le d_k}\; P_k,
\end{equation}
where \(P_k\) is the spectrum power, \(d_k\) is a small tolerance (in bins), and \(M_{\text{valid}}(\omega)\) counts the harmonics whose windows lie in-band (we require \(M_{\text{valid}}(\omega)\ge h_{\mathrm min}\)).

We observe that frequently, two overlapping rotors will project onto a single pixel (Figure~\ref{fig:main} (i)), creating power spectra composed with multiple fundamentals. Thus, instead of computing the SNR with the mean power from the complement set of bins, we utilize the median score to limit the secondary rotor's impact on the local noise floor.

Let $\mathcal{H}(\omega)$ be the set of valid harmonics of $\omega$,
\begin{equation}
\mathcal{H}(\omega) = \left\{\, P_k : k \in \bigcup_{m=1}^{M_{\text{valid}}(\omega)} [\,m\omega-d_k,\; m\omega+d_k\,] \right\},
\end{equation}
and $\overline{\mathcal{H}(\omega)}$ be the complement set of the power spectrum.
We form a robust, candidate-specific baseline using the median power outside the candidate’s harmonic windows $M(\overline{\mathcal{H}(\omega)})$.
The resulting median-normalized SNR is
\begin{equation}
\mathrm{SNR}_{M}(\omega) \;=\; \frac{N(\omega)}{M(\overline{\mathcal{H}(\omega))}},
\end{equation}
and pixels satisfying
\begin{equation}
\max_{\omega}\, \mathrm{SNR}_{M}(\omega) \;\ge\; \tau_{\text{comb}}
\end{equation}
are considered rotor pixels and their dominant frequencies are returned.

\begin{figure}[t]
    \centering
    \includegraphics[width=\linewidth]{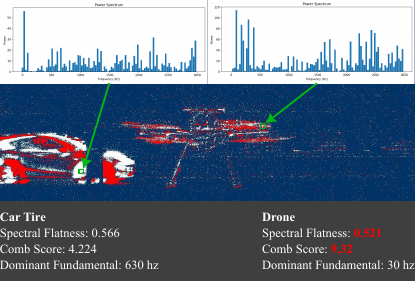}
    \caption{Visualization of one frame, comparing the spectra of a drone pixel versus a car wheel pixel. Both spectral flatness and harmonic comb scores from DDHF are able to distinguish between the spectrum of a spinning tire versus a spinning rotor.}
    \label{fig:car_vs_drone}
\end{figure}

\subsection{Algorithm Implementation}

\begin{figure}[t]
    \centering
    \includegraphics[width=\linewidth]{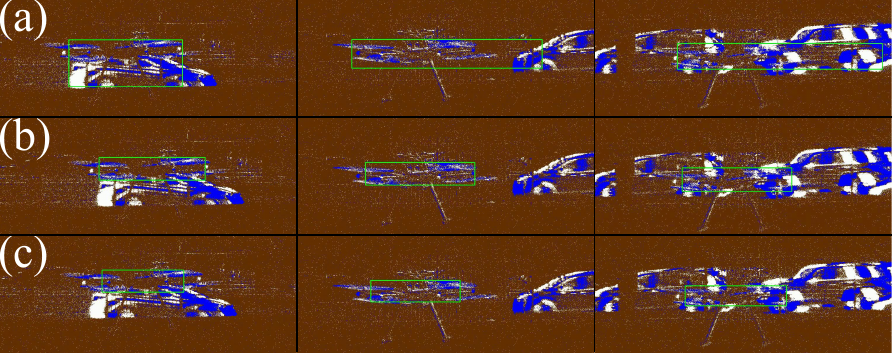}
    \caption{Demonstration of the interpretable tuning capabilities of DDHF. For row \textbf{(a)}, the default hyperparameters obtained from the Bayesian optimization were used. For row \textbf{(b)}, $\tau_{sf}$ was tuned from $0.89$ to $0.5$. For row \textbf{(c)}, $\tau_{\mathrm{comb}}$ was tuned from $1.5$ to $6$. In both tuning cases, the bounding box becomes far more selective, filtering out the power spectra of the moving cars and effectively isolating the drone. Interestingly, only the inner parts of the drone rotor at this pose appear to be identified.}
    \label{fig:tuning}
\end{figure}

\begin{table*}[t]
\centering
\small
\setlength{\tabcolsep}{4pt}
\begin{tabular}{l
    S[table-format=2.2] S[table-format=2.2] S[table-format=2.2]
    S[table-format=2.2] S[table-format=2.2] S[table-format=2.2]
    S[table-format=2.2] S[table-format=2.2] S[table-format=2.2]
    S[table-format=2.2] S[table-format=2.2] S[table-format=2.2]
    S[table-format=4.0]}
\toprule
\multicolumn{1}{c}{Condition} &
\multicolumn{3}{c}{Detection \textbf{Ours}} &
\multicolumn{3}{c}{Detection \textbf{YOLO}} &
\multicolumn{3}{c}{Latency (ms) \textbf{Ours}} &
\multicolumn{3}{c}{Latency (ms) \textbf{YOLO}} &
{n} \\
\cmidrule(lr){2-4}\cmidrule(lr){5-7}\cmidrule(lr){8-10}\cmidrule(lr){11-13}
& {P} & {R} & {F1}
& {P} & {R} & {F1}
& {Median} & {IQR} & {p95}
& {Median} & {IQR} & {p95} & {} \\
\midrule
Stationary - 10\,m
  & 99.58 & 99.58 & \textbf{99.58}
  & \multicolumn{1}{c}{\textit{99.79}} & \multicolumn{1}{c}{\textit{98.98}} & \multicolumn{1}{c}{\textit{99.38}}
    & \textbf{2.30} & 0.65 & 4.10
  & \multicolumn{1}{c}{\textit{11.91}} & \multicolumn{1}{c}{\textit{1.17}} & \multicolumn{1}{c}{\textit{15.24}}
  & 979 \\%
Slow - 10\,m
    & 97.18 & 96.57 & \textbf{96.87}
  & \multicolumn{1}{c}{\textit{92.60}} & \multicolumn{1}{c}{\textit{97.13}} & \multicolumn{1}{c}{\textit{94.81}}
  & \textbf{2.60} & 1.06 & 5.06 
  & \multicolumn{1}{c}{\textit{12.44}} & \multicolumn{1}{c}{\textit{1.18}} & \multicolumn{1}{c}{\textit{16.67}}
  & 350 \\%
Fast - 10\,m
  & 100.00 & 93.24 & 96.46
  & \multicolumn{1}{c}{\textit{97.26}} & \multicolumn{1}{c}{\textit{98.61}} & \multicolumn{1}{c}{\textbf{\textit{97.93}}}
   & \textbf{3.01} & 1.84 & 7.49
  & \multicolumn{1}{c}{\textit{12.07}} & \multicolumn{1}{c}{\textit{0.74}} & \multicolumn{1}{c}{\textit{13.45}}
  & 77 \\%

Multi-Drone - 15\,m
  & 93.63 & 93.53 & \textbf{93.58}
  & \multicolumn{1}{c}{\textit{100.00}} & \multicolumn{1}{c}{\textit{5.47}} & \multicolumn{1}{c}{\textit{10.37}}
  & \textbf{1.85} & 0.46 & 3.48
  & \multicolumn{1}{c}{\textit{11.57}} & \multicolumn{1}{c}{\textit{1.17}} & \multicolumn{1}{c}{\textit{14.69}}
  & 448 \\%
  
Direct Sunlight
  & 100.00 & 99.42 & \textbf{99.71}%
  & \multicolumn{1}{c}{\textit{100.00}} & \multicolumn{1}{c}{\textit{96.53}} & \multicolumn{1}{c}{\textit{98.24}}
  & \textbf{2.29} & 0.31 & 2.80%
  & \multicolumn{1}{c}{\textit{13.22}} & \multicolumn{1}{c}{\textit{1.26}} & \multicolumn{1}{c}{\textit{16.78}}
  & 173 \\%
  
Shaky Camera
  & 96.97 & 97.71 & \textbf{97.34}
  & \multicolumn{1}{c}{\textit{85.14}} & \multicolumn{1}{c}{\textit{96.18}} & \multicolumn{1}{c}{\textit{90.32}}
  & \textbf{2.57} & 0.37 & 3.99%
  & \multicolumn{1}{c}{\textit{13.70}} & \multicolumn{1}{c}{\textit{1.08}} & \multicolumn{1}{c}{\textit{14.85}}
  & 131 \\%
  
Moving Cars
  & 47.06 & 47.06 & \textbf{47.06}
  & \multicolumn{1}{c}{\textit{0.00}} & \multicolumn{1}{c}{\textit{0.00}} & \multicolumn{1}{c}{\textit{0.00}}
  & \textbf{3.30} & 0.41 & 4.65%
  & \multicolumn{1}{c}{\textit{13.70}} & \multicolumn{1}{c}{\textit{1.08}} & \multicolumn{1}{c}{\textit{14.85}}
  & 272 \\%
  
Far Distance - 300m
  & 77.04 & 69.33 & \textbf{72.98}
  & \multicolumn{1}{c}{\textit{0.00}} & \multicolumn{1}{c}{\textit{0.00}} & \multicolumn{1}{c}{\textit{0.00}}
   & \textbf{2.10} & 0.52 & 3.06
  & \multicolumn{1}{c}{\textit{13.70}} & \multicolumn{1}{c}{\textit{1.08}} & \multicolumn{1}{c}{\textit{14.85}}
  & 150 \\%

  \midrule
  
Weighted Averages
  & 91.28 & 90.53 & \textbf{90.89}
  & \multicolumn{1}{c}{\textit{81.72}} & \multicolumn{1}{c}{\textit{65.98}} & \multicolumn{1}{c}{\textit{66.47}}
   & \textbf{2.39} & 0.64 & 4.13
  & \multicolumn{1}{c}{\textit{12.40}} & \multicolumn{1}{c}{\textit{1.15}} & \multicolumn{1}{c}{\textit{15.30}}
  & 2580 \\%

\bottomrule
\end{tabular}
\caption{Head-to-head detection and latency at IoU=0.5. Our method is shown to almost always achieve far higher average detection scores and significantly lower latencies. As each frame at 30 fps corresponds to 33.33 milliseconds, latencies are comfortably below this budget, indicating our method is likely sufficient for real-time operation on the tested hardware.
Latency is measured in milliseconds per frame, and inference is performed on an NVIDIA L40-8Q GPU. $n$ = number of frames tested. We also report metric averages, weighted by number of frames per experiment.}
\label{tab:comparison}
\end{table*}

\noindent\textbf{Event Processing}
Events are streamed in temporal windows of 33.3 ms (30 FPS) to make use of GPU parallelism. For each window, events are moved to the GPU, sorted by pixel index, and processed with our algorithm to extract dominant fundamental frequencies. Pixels that accumulate fewer than $n_{\min}$ events within the window are discarded to avoid degenerate transforms, where $n_{\min}$ is selected empirically (details in Section~\ref{sec:exp_setup}).

For each active pixel, we extract polarity and timestamp arrays and directly compute the NDFT. The polarity stream is mapped to ${-1, +1}$ and timestamps are normalized to $[-\pi, \pi)$. We evaluate integer frequency modes $k = 0,\dots,K - 1$, compute power $P_k = |F_k|^2$, and filter candidate pixels via spectral flatness. Remaining responses are scored using the harmonic comb score threshold, producing a boolean rotor segmentation mask, where true values represent a rotor detection.
\\\\
\noindent\textbf{Segmentation Mask Processing}
We apply 8-connected component labeling to the boolean mask to form initial rotor sections, and we extract their bounding boxes. Small boxes below the area threshold $A_{\min}$ are removed. We observe a tendency for bounding boxes to surround individual rotors, producing multiple boxes per drone. Thus, boxes are merged whose centroids are sufficiently close relative the rotor's size. Specifically, two boxes with widths and heights $(w_i,h_i)$ and centroids $c_i$ are merged if
\begin{align}
    \mathrm{dist}(c_1, c_2) < \alpha \cdot \max(w_1, h_1, w_2, h_2),
\end{align}
where $\alpha$ is a scale factor tuned on validation data. We use the maximum box dimension as scale, as for example the rotor width is stable even when the drone tilts towards the camera while the height collapses. The merged boxes form final drone detections.

This merge rule follows from the fact that, under an ideal pinhole camera, objects at similar depth shrink uniformly as they move away, preserving the ratio between their projected lengths and the distances between them. Using the rotor length (the longer box dimension) as the scale therefore ensures that boxes from the same drone are consistently merged across different depths and viewing angles.

\section{Results}
\label{sec:results}

\subsection{Experimental Setup}
\label{sec:exp_setup}
\noindent\textbf{Data collection.}
To simulate real-world deployment, we capture test and training data separately across two different days, environments, and scenarios. The training set consists of four short videos of the DJI Mavic Pro (about 5 seconds each): a slow speed traversal in front of a cluttered background and a sweep over at stationary, medium, and fast-pass speeds with the sky as background. The test set consists of a sweep of drones and speeds: 9 different short videos, taken across three speeds ($0 m/s$, $1\text{-}2 m/s$, and $6\text{-}7 m/s$) and three different drones. Furthermore, we collect an additional dataset of miscellaneous conditions to address potential edge cases as well as highlight the benefits of event cameras. We include a multi-drone test with two stationary drones at $\approx\!30m$ to test the box merging step, a slow pass over the focal plane with the camera facing into direct sunlight to demonstrate the large dynamic range of event cameras, a handheld shaky camera that tests whether our algorithm simply searches for active pixels or truly identifies a rotor fingerprint, and a stationary drone hovering in front of a busy street, where the spinning tires of the car may confuse the algorithm. Finally, we collect a third dataset, varying the angle of elevation of the drone to the camera from $0^\circ$ to $90^\circ$ in $10^\circ$ increments to investigate how the pose of the rotors affects the accuracy of our algorithm.

We hand-label bounding boxes that circumscribe all rotors of the drone, which includes the entire body of the drone for most poses. We choose to use only a small validation set to showcase our method's generalization ability and applicability to real-world scenarios where data may be limited. Drones utilized include the DJI Mavic Pro, Autel Evo 1, Autel Evo 2, and Inspired Flight IF1600A drones for the test data. For all collections, we use the EVK4-HD event camera from Prophesee, which features the IMX636 1280x720 pixel CMOS vision sensor. For the moving cars test, we utilize the Navitar 100mm F2.8 lens, and for the angle of elevation experiment, the Navitar 35mm F1.4 is used. The remaining experiments utilize the kit 8mm F2.0 lens from Prophesee.
\\\\
\noindent\textbf{Hyperparameter Search.}
We tune thresholding and aggregation hyperparameters with Bayesian optimization (BO) on a small validation set of three hand-labeled videos. BO uses a Gaussian–process surrogate with expected improvement and uniform priors over the ranges in Table~\ref{tab:bo-space}. The bounds are chosen from physical constraints of drone rotors and empirical distributions measured on training clips. For the search, we use 50 random evaluation points, then perform BO for 300 iterations.

\begin{figure}[t]
  \centering
  \begin{subfigure}[t]{0.8\linewidth}
    \centering
    \includegraphics[width=\linewidth]{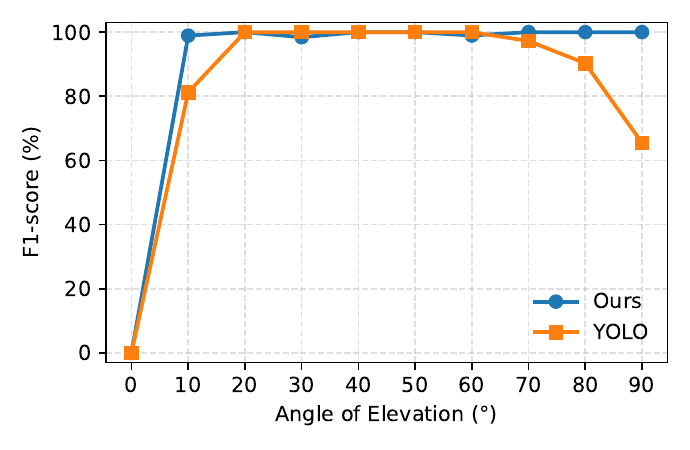}
    \caption{F1-score vs. angle of elevation.}
    \label{fig:angle_results_f1}
  \end{subfigure}
  \\
  \begin{subfigure}[t]{0.8\linewidth}
    \centering
    \includegraphics[width=\linewidth]{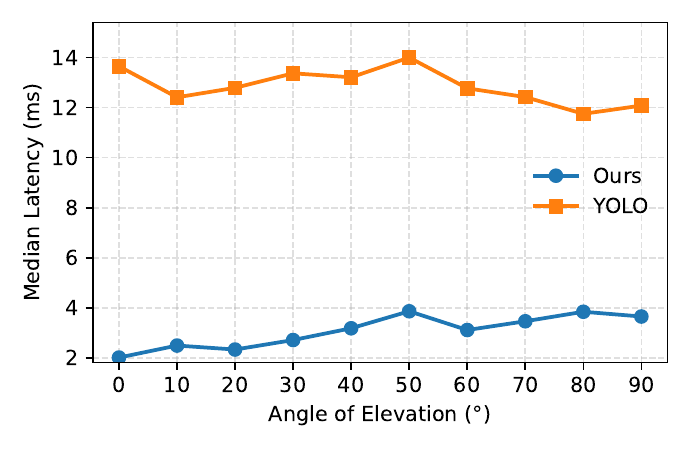}
    \caption{Median latency vs. angle of elevation.}
    \label{fig:angle_results_latency}
  \end{subfigure}
  \caption{Performance versus angle of elevation. (a) Accuracy measured by F1-score; (b) inference latency.}
  \label{fig:angle_results}
\end{figure}

We run BO for $T$ iterations with the same validation split used for the earlier grid search, optimizing F1@0.5 IoU. The best setting selected by BO is later \emph{fixed} for all test results.
\\\\
\noindent\textbf{YOLO Comparison.} To benchmark our non-neural, physics-guided detection pipeline against a widely adopted deep learning model, we train and evaluate a YOLO baseline under identical experimental conditions. Because YOLO operates on image inputs, we convert the event stream into accumulated frames across 33.3ms windows. YOLO is chosen because it represents the de facto standard in real-time object detection, offering a strong reference point for both performance and efficiency. This comparison is particularly relevant because, despite YOLO’s success, purely data-driven models can struggle with generalization outside their training distribution—especially in low-data or domain-shifted scenarios such as UAV detection under varying viewpoints, illumination, and motion blur. By contrast, our non-neural approach leverages domain knowledge and signal-level priors to achieve robustness without relying on extensive retraining or large labeled datasets.

For the YOLO baseline, we use the lightweight YOLOv11-n architecture, which contains 2.6 million trainable parameters. The original validation dataset is divided into an 80/20 train/validation split, while testing is performed on the same held-out test set used by our algorithm to ensure fair comparison. We conduct a hyperparameter sweep over 40 iterations of 2 epochs each, then retrain the model using the best-performing configuration. With early stopping, the model converges after 19 epochs, reaching a validation precision of 99.33\% at 0.5 IoU.

During inference, we retain detections whose confidence exceeds 0.3, corresponding to the threshold that achieved the best F1 score on the validation set. Quantitative results are summarized in Table~\ref{tab:comparison} and Figure~\ref{fig:angle_results}, demonstrating that our non-neural approach attains competitive accuracy and significantly lower latency despite its model-free design.

\subsection{Findings}

We evaluate DDHF across a set of experiments designed to test accuracy, latency, and robustness under diverse real-world conditions. Metrics are reported at 0.5 IoU using the held-out test set described in Section~\ref{sec:exp_setup}. Results across a miscellaneous test suite are summarized in Table~\ref{tab:comparison} and angle of elevation experimental results are shown in Figure~\ref{fig:angle_results}. Overall, DDHF delivers consistently high detection accuracy and maintains real-time performance across challenging lighting, motion, and multi-object scenarios. Its analytic formulation provides predictable, physically interpretable failure modes—primarily tied to optical focus and rotor pose—rather than data-driven variance, distinguishing it from neural baselines.

\noindent\textbf{Core Performance and Latency.} 
Across stationary, slow, and fast passes, DDHF achieves comparable or superior F1 scores to YOLO, while maintaining sub-frame latencies well below the 33.3~ms frame budget. Stationary and slow-speed trials show a clear advantage for DDHF, while YOLO slightly outperforms at the fastest traversal. These results confirm that per-pixel harmonic fingerprinting remains stable across a broad range of motion dynamics.

\noindent\textbf{Multi-Drone and Scene Complexity.}
In the multi-drone test, DDHF achieves an F1 of 93.58\%, correctly identifying both UAVs while preserving separate bounding boxes. This reflects successful hyperparameter selection of $\alpha$ for rotor-aware box merging. In contrast, YOLO’s performance drops sharply, likely due to the out-of-distribution nature of the scene.

\noindent\textbf{Illumination Robustness.}
Under direct sunlight, DDHF highlights the dynamic range advantage of event-based sensors: both our method and YOLO achieve high F1 scores. This result illustrates the resilence of temporal contrast sensing to extreme illumination, providing sharp drone resolution in a scenario where conventional CMOS sensors would likely saturate.

\noindent\textbf{Camera Motion and Background Clutter.}
In the handheld shaky-camera sequence, DDHF sustains high accuracy despite dense background activity, confirming the robustness of the rotor fingerprinting method against ego-motion and high event throughput. In the moving-cars test, performance degrades moderately due to wheel spokes and fences producing pseudo-periodic signals, yet DDHF remains stronger than YOLO, which fails completely. But, we perform post-test tuning of both the spectral flatness threshold $\tau_{sf}$ and the rotor-comb threshold $\tau_{\text{comb}}$ further suppresses these false positives, demonstrating DDHF’s interpretability and ad-hoc tunability (see Fig.~\ref{fig:tuning}).

\noindent\textbf{Long-Range and Pose Sensitivity.}
At long range, DDHF detects drones up to 300~m using a 100~mm F2.8 lens, but accuracy decreases at the lower angle of elevation ($0^\circ$--$10^\circ$) where rotors become edge-on to the sensor \footnote{However, in all frames, the individual drone rotors are detected. The decrease in accuracy is due to the box merging step, which sometimes fails to merge each rotor together. This behavior may be due to the lower angle of elevation causing our method to reject outer rotor pixels, producing smaller bounding boxes which fall beneath the $\alpha$ merge hyperparameter.}. This geometric limitation aligns with the trends shown in Fig.~\ref{fig:angle_results}, where performance for both methods drops at $0^\circ$, while DDHF maintains stable accuracy across mid and high elevations, unlike YOLO, which degrades above $70^\circ$ due to limited training diversity.

\begin{table}[h]
\centering
\small
\begin{tabular}{lccc}
\toprule
Parameter & Symbol & BO Range & Final Value \\
\midrule
Spectral flatness threshold & $\tau_{sf}$ & $[0.3, 0.98]$ & 0.89 \\
Comb SNR threshold & $\tau_{\mathrm{comb}}$ & $[1.5, 6.0]$ & 1.5 \\
\# comb harmonics & $M$ & $[4, 16]$ & 4 \\
Harmonic half-bandwidth & $\Delta k$ & $[0, 4]$ & 3 \\
Min. matched harmonics & $h_{\min}$ & $[2, 8]$ & 4 \\
Minimum passes / window & $n_{\min}$ & $[2, 10]$ & 6 \\
Minimum box area (px) & $A_{\min}$ & $[9, 196]$ & 147 \\
Join radius factor & $\alpha$ & $[3.0, 20.0]$ & 19.05 \\
\bottomrule
\end{tabular}
\caption{Bayesian optimization search space and results. All hyperparameters saw Uniform priors, and we use a GP with the expected improvement (EI) acquisition function.}
\label{tab:bo-space}
\end{table}

\section{Conclusion}
\label{sec:conclusion}

We introduce DDHF, a fully analytic framework for real-time UAV detection from event camera data. By leveraging the NDFT, our method directly extracts harmonic frequency signatures from asynchronous event streams without approximating a uniformly-sampled DFT. This approach preserves the fine-grained temporal fidelity of event cameras while providing interpretable, tunable frequency-domain fingerprinting.

Across a diverse evaluation suite, including challenging conditions such as direct sunlight, motion blur, and multi-drone scenes, DDHF consistently achieves higher accuracy and significantly lower latency than a YOLOv11-n baseline trained under identical conditions. Our per-pixel frequency analysis reaches an average F1 score of 90.9\% with 2.39 ms per-frame latency, outperforming YOLO’s 66.7\% F1 and 12.4 ms latency. These results demonstrate that domain-driven, signal-level approaches can surpass deep learning methods, particularly in low-data or domain-shifted settings.

Beyond performance, DDHF offers a high degree of interpretability and adaptability as parameters can be intuitively tuned to balance selectivity and recall for different environments, allowing for rapid deployment without retraining.

However, DDHF’s accuracy is still influenced by optical and geometric constraints. Detection performance decreases at low angles of elevation, where rotor planes align with the image sensor and become difficult to distinguish. Similarly, achieving long-range detection requires higher magnification optics---for example, our 300 m tests depend on a 100 mm focal length lens---whereas radar-based systems naturally maintain range independence. These factors highlight that event-based vision, while precise and fast, remains bounded by optical focus and viewing geometry.

Future directions include extending this framework toward on-drone, edge-based UAV detection, testing different power spectra fingerprinting techniques, and event-radar fusion, enabling scalable, physics-grounded perception systems.

{\small
\bibliographystyle{ieeetr}
\bibliography{PaperForReview}
}
\end{document}